\documentclass[fleqn,10pt]{wlscirep}
\usepackage[utf8]{inputenc}
\usepackage[T1]{fontenc}
\usepackage{color}
\usepackage{caption}
\usepackage{subfig}

\usepackage{newtxtext,newtxmath}
\usepackage{cases}

\renewcommand{\vec}{\boldsymbol}

 \graphicspath{{Figures/}}

\usepackage[nolist]{acronym}
\title{Echo State Networks with Self-Normalizing Activations on the Hyper-Sphere}

\author[1,*]{Pietro Verzelli}
\author[1,2]{Cesare Alippi}
\author[3,4]{Lorenzo Livi}
\affil[1]{Faculty of Informatics, Università della Svizzera Italiana , Lugano, 69000, Switzerland.}
\affil[2]{Department of Electronics, Information and bioengineering, Politecnico di Milano, Milan, 20133, Italy.}
\affil[3]{Departments of Computer Science and Mathematics, University of Manitoba, Winnipeg, MB R3T 2N2, Canada.}
\affil[4]{Department of Computer Science, College of Engineering, Mathematics and Physical Sciences, University of Exeter, Exeter EX4 4QF, United Kingdom.}

\affil[*]{corresponding author(mail: pietro.verzelli@usi.ch)}

\begin{abstract}
Among the various architectures of Recurrent Neural Networks, Echo State Networks (ESNs) emerged due to their simplified and inexpensive training procedure. These networks are known to be sensitive to the setting of hyper-parameters, which critically affect their behavior. Results show that their performance is usually maximized in a narrow region of hyper-parameter space called edge of criticality. Finding such a region requires searching in hyper-parameter space in a sensible way: hyper-parameter configurations marginally outside such a region might yield networks exhibiting fully developed chaos, hence producing unreliable computations. The performance gain due to optimizing hyper-parameters can be studied by considering the memory--nonlinearity trade-off, i.e., the fact that increasing the nonlinear behavior of the network degrades its ability to remember past inputs, and vice-versa.
In this paper, we propose a model of ESNs that eliminates critical dependence on hyper-parameters, resulting in networks that provably cannot enter a chaotic regime and, at the same time, denotes nonlinear behavior in phase space characterized by a large memory of past inputs, comparable to the one of linear networks. Our contribution is supported by experiments corroborating our theoretical findings, showing that the proposed model displays dynamics that are rich-enough to approximate many common nonlinear systems used for benchmarking.
\end{abstract}
\begin{document}

\flushbottom
\maketitle
%
%
\thispagestyle{empty}

\section*{Introduction}

Although the use of \emph{\acp{RNN}} in machine learning is boosting, also as effective building blocks for deep learning architectures, a comprehensive understanding of their working principles is still missing \cite{sussillo2013opening,ceni-ff-18}.
Of particular relevance are \acp{ESN}, introduced by Jaeger \cite{jaeger2004harnessing} and independently by Maass \emph{et al.}\cite{maass2002real} under the name of \ac{LSM}, which emerge from \acp{RNN} due to their training simplicity.
The basic idea behind \acp{ESN} is to create a randomly connected recurrent network, called reservoir, and feed it with a signal so that the network will encode the underlying dynamics in its internal states. The desired -- task dependent -- output is then generated by a readout layer (usually linear) trained to match the states with the desired outputs. Despite the simplified training protocol, \acp{ESN} are universal function approximators \cite{grigoryeva2018echo} and have shown to be effective in many relevant tasks \cite{pathak2017using, pathak2018model, pathak2018hybrid, bianchi2015prediction,bianchi2018reservoir, palumbo2016human, gallicchio2018comparison}.

These networks are known to be sensitive to the setting of hyper-parameters like the \ac{SR}, the input scaling and the sparseness degree \cite{jaeger2004harnessing}, which critically affect their behavior and, hence, the performance at task.
Fine tuning of hyper-parameters requires cross-validation or ad-hoc criteria for selecting the best-performing configuration. Experimental evidence and some results from the theory show that \acp{ESN} performance is usually maximized in correspondence of a very narrow region in hyper-parameter space called \ac{EoC} \cite{sompolinsky1988chaos,esnfish2016, verzelli2018characterization, legenstein2007edge,bertschinger2004real,rajan2010stimulus, rivkind2017local, gallicchio2018chasing}.
However, we comment that beyond such a region \acp{ESN} behave chaotically, resulting in useless and unreliable computations. 
At the same time, it is everything but trivial configuring the hyper-parameters to lie on the EoC still granting a non-chaotic behavior.
A very important property for \acp{ESN} is the \ac{ESP}, which basically asserts that their behavior should depend on the signal driving the network only, regardless of its initial conditions \cite{yildiz2012re}.
Despite being at the foundation of theoretical results \cite{grigoryeva2018echo}, the \ac{ESP} in its original formulation raises some issues, mainly because it does not account for multi-stability and is not tightly linked with properties of the specific input signal driving the network \cite{yildiz2012re,manjunath2013echo,wainrib2016local}.

In this context, the analysis of the memory capacity (as measured by the ability of the network to reconstruct or remember past inputs) of input-driven systems plays a fundamental role in the study of \acp{ESN}~\cite{tivno2013short, goudarzi2016memory, jaeger2002short, ganguli2008memory}.
In particular, it is known that \acp{ESN} are characterized by a memory--nonlinearity trade-off \cite{dambre2012information, verstraeten2010memory, inubushi2017reservoir}, in the sense that introducing nonlinear dynamics in the network degrades memory capacity.
Moreover, it has been recently shown that optimizing memory capacity does not necessarily lead to networks with higher prediction performance \cite{marzen2017difference}.

In this paper, we propose an \ac{ESN} model that eliminates critical dependence on hyper-parameters, resulting in models that cannot enter a chaotic regime. In addition to this major outcome, such networks denote nonlinear behavior in phase space characterized by a large memory of past inputs: the proposed model generates dynamics that are rich-enough to approximate nonlinear systems typically used as benchmarks.
Our contribution is based on a nonlinear activation function that normalizes neuron activations on a hyper-sphere.
We show that the spectral radius of the reservoir, which is the most important hyper-parameter for controlling the \ac{ESN} behavior, plays a marginal role in influencing the stability of the proposed model, although it has an impact on the capability of the network to memorize past inputs. Our theoretical analysis demonstrates that this property derives from the impossibility for the system to display a chaotic behavior: in fact, the maximum Lyapunov exponent is always null. An interpretation of this very important outcome is that the network always operates on the \ac{EoC}, regardless of the setting chosen for its hyper-parameters.
We tested the memory of our model on a series of benchmark time-series, showing that its performance for memorization tasks is comparable with that exposed by linear networks. We also explored the memory--nonlinearity trade--off following \cite{inubushi2017reservoir}, showing that our model outperforms networks with linear and hyperbolical activations when both memory and nonlinearity are required by the task at hand. 

\section*{Results}

\subsection*{Echo state networks}

An \ac{ESN} without output feedback connections is defined as:
\begin{subequations} 
\label{eqn:ESN}
    \begin{align}
    &\vec{a}_k = W \vec{x}_{k-1} + W_{\text{in}}\vec{s}_{k} \label{eqn:linear_update}\\
    &\vec{x}_{k} = \phi(\vec{a}_k) \label{eqn:activation}\\
    &\vec{y}_k = W_\text{out}\vec{x}_k\label{eqn:reservoir_readout}
    \end{align}
\end{subequations}
where $\vec{x}_k \in \mathbb{R}^{N}$ is the system state at time-step $k$, $N$ is the number of hidden neurons composing the network reservoir, and $W \in \mathbb{R}^{N \times N}$ is the connection matrix of the reservoir. 
The signal value at time $k$, $\vec{s}_k \in \mathbb{R}^{N_\text{in}}$, is processed by the input-to-reservoir matrix $W_{\text{in}} \in \mathbb{R}^{N \times N_\text{in}}$. 
The activation function $\phi$ usually takes the form of the hyperbolic tangent function, for which the network is a universal function approximator~\cite{grigoryeva2018echo}.
Also linear networks (i.e., when $\phi$ is the identity) are commonly studied, both for the proven impact on applications and the very interesting results that can be derived in closed-form\cite{tivno2013short, goudarzi2016memory, marzen2017difference, ganguli2008memory, tivno2018asymptotic}.
Other activation functions have been proposed in the neural networks literature, including those that normalize the activation values on a closed hyper-surface\cite{andrecut2017reservoir} and those based on non-parametric estimation via composition of kernel functions\cite{scardapane2019kafnets}.

\begin{figure}
    \centering
    \includegraphics[width = 0.7\textwidth]{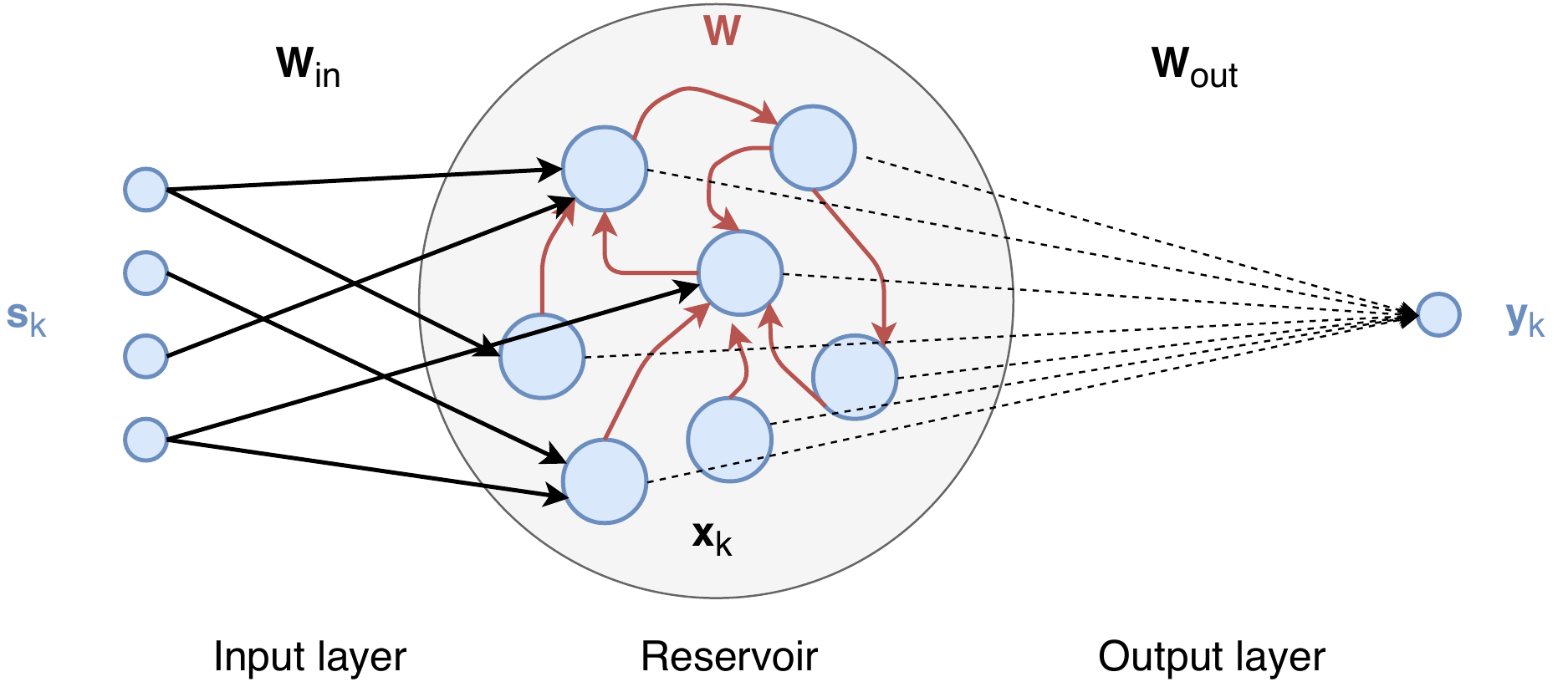}
    \caption{Schematic representation of an \ac{ESN} with a $4$-dimensional input, a mono-dimensional output and $N = 7$ neurons in the reservoir. Fixed connections are drawn using continuous lines while dashed lines represent connections that are learned. }
    \label{fig:ESN}
\end{figure}

The output $\vec{y} \in \mathbb{R}^{N_\text{out}}$ is generated by the matrix $W_\text{out}$, whose weights are learned: generally by ridge regression or lasso\cite{jaeger2004harnessing, lukovsevivcius2009reservoir} but also with online training mechanisms\cite{sussillo2009generating}. In \acp{ESN}, the training phase does not affect the dynamics of the system, which are \emph{de facto} task-independent. 
It follows that once a suitable reservoir rich in dynamics is generated, the same reservoir may serve to perform different tasks.
A schematic representation of and \ac{ESN} in depicted in Fig. \ref{fig:ESN}

\subsection*{Self-normalizing activation function}
Here, we propose a new model for \acp{ESN} characterized by the use of a particular self-normalizing activation function that provides important features to the resulting network.
Notably, the proposed activation function allows the network to exhibit nonlinear behaviors and, at the same time, provides memory properties similar to those observed for linear networks. 
This superior memory capacity is linked to the fact that the network never displays a chaotic behavior: we will show that the maximum Lyapunov exponent is always zero, implying that the network operates on the \ac{EoC}.
The proposed activation function guarantees that the \ac{SR} of the reservoir matrix (whose value is used as a control parameter) can vary in a wide range without affecting the network stability.

\begin{figure}[ht]
\centering
\includegraphics[width=0.5 \linewidth]{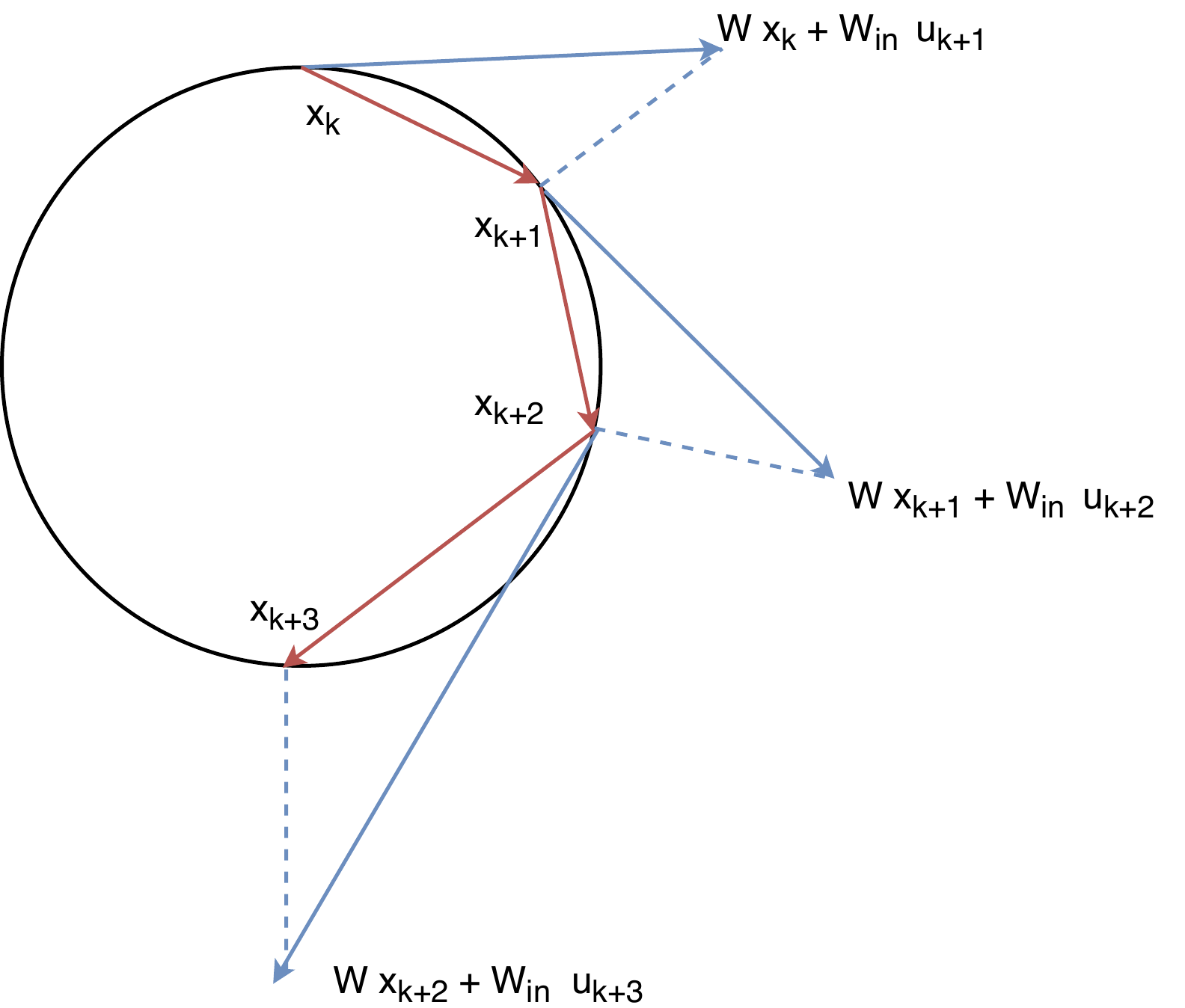}
\caption{Example of the behavior of the proposed model in a $2$-dimensional scenario. The blue lines represent the linear update step of Eq. \eqref{eqn:new_linear_update}, while the dashed lines denote the projection of Eq. \eqref{eqn:new_activation}. The red lines represent the actual steps performed by the system.
Note that condition \eqref{eqn:condition} accounts for the fact that the linear step must never bring the system state inside the hyper-sphere.}
\label{fig:diag}
\end{figure}

The proposed self-normalizing activation function is
\begin{subequations} 
\label{eqn:new_system}
    \begin{align}
    &\vec{a}_k = W \vec{x}_{k-1} + W_{\text{in}}\vec{s}_{k} \label{eqn:new_linear_update}\\
    &\vec{x}_{k} = r \frac{\vec{a}_k}{\lVert \vec{a}_k\rVert} \label{eqn:new_activation}
    \end{align}
\end{subequations}
and leaves the readout \eqref{eqn:reservoir_readout} unaltered.
The normalization in Eq.~\ref{eqn:new_activation} projects the network pre-activation $a_k$ onto an $(N-1)$-dimensional hyper-sphere $\mathbb{S}^{N-1}_{r} := \{ \vec{p} \in \mathbb{R}^{N},  \lVert \vec{p} \rVert = r \}$ of radius $r$.  
Fig.~\ref{fig:diag} illustrates the normalization operator applied to the state.
Note that the operation \eqref{eqn:new_activation} is not element-wise like most of activation functions as its effect is global, meaning that a neuron's activation value depends on all other values.

%
%
%
%

\subsection*{Universal function approximation property}
The fact that recurrent neural networks are universal function approximators has been proven in previous works \cite{siegelmann1995computational, hammer2000approximation} and some results on the universality of reservoir-based computation are given in\cite{maass2002real,hammer2003recurrent}. Recently, Grigoryeva and Ortega \cite{grigoryeva2018echo} proved that \acp{ESN} share the same property.
Here, we show that the universal function approximation property also holds for the proposed \acp{ESN} model \eqref{eqn:new_system}. 
To this end, we define a \emph{squashing function} as a map $f: \mathbb{R} \to [-1,1]$ that is non decreasing and saturating, i.e., $\lim_{x \to \pm \infty} f(x) = \pm 1$.
We note that $\phi_i(\vec{x}) := x_i/ \lVert \vec{x} \rVert$ can be intended as a squashing function for each $i$th component. 
In the same work\cite{grigoryeva2018echo}, the authors show that an \ac{ESN} in the form \eqref{eqn:ESN} is a universal function approximator provided it satisfies the \emph{contractivity condition}: $\lVert \phi(x_1) - \phi(x_2) \rVert \le \lVert x_1 - x_2 \rVert$ for each $x_1$ and $x_2$, when $\phi$ is a squashing function.
Jaeger and Haas\cite{jaeger2004harnessing} showed that this condition is sufficient to grant the \ac{ESP}, implying that \acp{ESN} are universal approximators.

We prove in the Methods section that \eqref{eqn:new_system} grants the contractivity condition, provided that:
\begin{equation}\label{eqn:condition}
    \sigma_\text{min}(W) \ge 1 + \frac{\lVert W_\text{in} \rVert \lVert\vec{s}_{\text{max}}\rVert}{r}
\end{equation}
where $\sigma_\text{min}(W)$ is the smallest singular value of matrix $W$ and $\lVert\vec{s}_{\text{max}}\rVert$ denotes the largest norm associated to an input.
Eq. \ref{eqn:condition} results in a sufficient yet not necessary condition that may be understood as requiring that the input will never be strong enough to contrast the expansive dynamics of the system, leading the network state inside the hyper-sphere of radius $r$.
In fact, unless the signal is explicitly designed for violating such a condition, it will very likely not bring the system inside the hyper-sphere as long as the norm of $W$ is large enough compared to the signal variance.

\subsection*{Network state dynamics: the autonomous case}

We now discuss the network state dynamics in the autonomous case, i.e., in the absence of input.
This allows us to prove why the network cannot be chaotic.

From now on, we assume $r=1$ as this does not affect the dynamics, provided that condition \eqref{eqn:condition} is satisfied. 
From \eqref{eqn:new_system}, the system state dynamics in the autonomous case reads:
\begin{equation}\label{enq:map}
     \phi(\vec{x}) := \frac{W\vec{x}}{\lVert W \vec{x} \rVert}
\end{equation}
At time-step $n$ the system state is given by
\begin{equation}\label{eqn:evolution}
    \vec{x}_{n} =  \phi(\vec{x}_{n-1}) = \frac{W\vec{x}_{n-1}}{\lVert W\vec{x}_{n-1} \rVert} = 
    \frac{W}{\lVert W W\frac{\vec{x}_{n-2}}{\lVert W \vec{x}_{n-2} \rVert} \rVert} 
    \frac{W\vec{x}_{n-2}}{\lVert W \vec{x}_{n-2} \rVert} =
    \frac{W^2\vec{x}_{n-2}}{\lVert W^2 \vec{x}_{n-2}\rVert}
\end{equation}
By iterating this procedure, one obtains:
\begin{equation}\label{eqn:evolution_IC}
    \vec{x}_{n} := \phi^{n}(\vec{x}_0)
    =\frac{W^{n}\vec{x}_{0}}{\lVert W^{n} \vec{x}_{0}\rVert}
\end{equation}
where $\vec{x}_0$ is the initial state. 
This implies that, for the autonomous case, a system evolving for $n$ time-steps coincides with updating the state by performing $n$ matrix multiplications and projecting the outcome only at the end.

It is worth to comment that this holds also if matrix $W$ changes over time.
In fact, let $W_n:=W(n)$ be $W$ at time time $n$. Then, the evolution of the dynamical system reads:
\begin{equation}\label{eqn:evolution_general}
    \vec{x}_{n} := \phi^n(\vec{x}_0)
    =\frac{W_n W_{n-1} \dots W_2 W_1 \vec{x}_{0}}{\lVert W_n W_{n-1} \dots W_2 W_1 \vec{x}_{0}\rVert}
\end{equation}
Furthermore, note that a system described by matrix $W$ and a system characterized by $W' = aW$ coincide. In turn, this implies that the \ac{SR} of the matrix does not alter the dynamics in the autonomous case.

\subsection*{Edge of criticality}

When tuning the hyper-parameters of \acp{ESN}, one usually tries to bring the system close to the \ac{EoC}, since it is in that region that their performance is optimal\cite{esnfish2016}.
This can be explained by the fact that, when operating in that regime, the system introduces rich dynamics without denoting chaotic behavior.

Here, we show that the proposed recurrent model \eqref{eqn:new_system} cannot enter a chaotic regime.
Notably, we prove that, when the number of neurons in the network is large, the maximum (local) Lyapunov exponent cannot be positive, hence neglecting the possibility to introduce a chaotic behavior.
To this end, we determine the Jacobian matrix of \eqref{eqn:new_activation} and then show that, since its spectral radius tends to $1$, the maximum \ac{LLE} must be null.
The Jacobian matrix of \eqref{eqn:new_activation} reads:
\begin{align} \label{eqn:jacobian}
    J_{ij} =
    \frac{\partial}{\partial x_i}\phi_j(\vec{x})
    &=
    \sum_l^N
    \frac{\partial \phi_j}{\partial a_l} \frac{\partial a_l}{\partial x_i}
    = 
    \frac{W_{ij}}{\lVert \vec{a} \rVert}\left( 1 - \frac{a_i a_j}{\lVert \vec{a} \rVert^2} \right)
\end{align}
where the time index $k$ is omitted to ease the notation.
We observe that, asymptotically for large networks ($N \to \infty$), we have that $a_i/ \lVert \vec{a} \rVert \to 0$ for each $i$, meaning that the Jacobian matrix reduces to $J(\vec{x}) = W / \lVert W\vec{x}\rVert$. As we are considering the case with $r=1$, we know that $\lVert \vec{x}\rVert = 1$. 

This allows us to approximate the norm of $W$ with its \ac{SR} $\rho = \rho(W)$,
\begin{equation}
\label{eqn:jacobian_approx}
    J 
    \approx 
    \frac{W}{\lVert W\vec{x}\rVert} 
    \approx 
    \frac{W}{\rho}.
\end{equation}
Under this approximation \eqref{eqn:jacobian_approx}, the largest eigenvalue of $J$ must be $1$ as the \ac{SR} $\rho$ is the largest absolute value among the eigenvalues of $W$.
We thus characterize the global behavior of \eqref{eqn:new_activation} by considering the maximum \ac{LLE}\cite{esnfish2016}, which is defined as:
\begin{equation}
\label{eqn:LLE}
    \Lambda := \lim_{n \to \infty}\frac{1}{n} \log \left( \prod_k^n \rho_k \right) 
\end{equation}
where $\rho_k$ is the spectral radius of the Jacobian at time-step $k$.
Eq.~\ref{eqn:LLE} implies that $\Lambda = 0$ as $n\rightarrow \infty$, hence proving our claim.

\begin{figure}[ht]
\centering
\includegraphics[width = \linewidth]{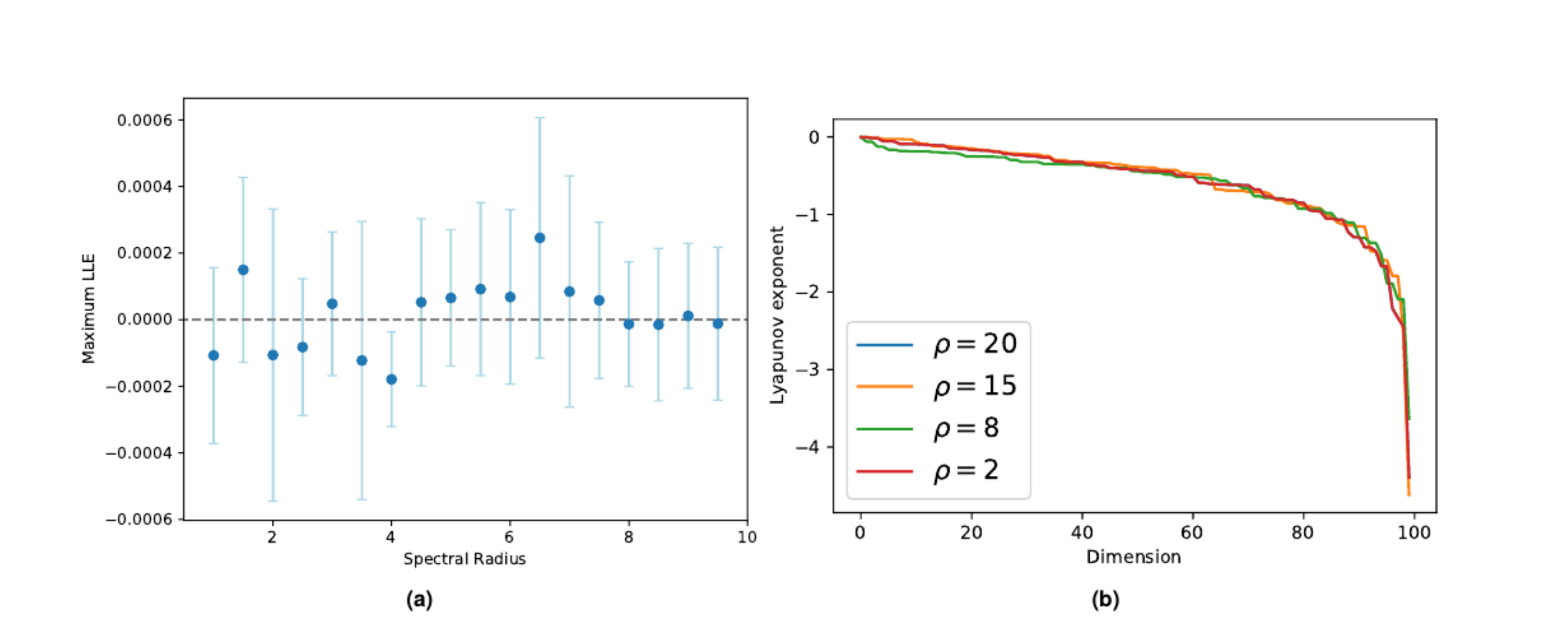}
\caption{Panel (a) shows \acp{LLE} for different value of the \ac{SR}. Each point in the plot represents the mean of $10$ different realizations, using a network with $N = 500$ neurons. 
In panel (b) examples of the Lyapunov spectrum for different $100$-dimensional networks are plotted. The Lyapunov exponents are ordered by decreasing magnitude.
Note that the largest exponent is always zero.}
\label{fig:Lyapunov}
\end{figure}

In order to demonstrate that $\Lambda = 0$ holds also for networks with a finite number $N$ of neurons in the recurrent layer, we numerically compute the maximum \ac{LLE} by considering the Jacobian in \eqref{eqn:jacobian}. The results are displayed in Fig.~\ref{fig:Lyapunov}.
Fig.~\ref{fig:Lyapunov} panel (a) shows the average value of the maximum \ac{LLE} with the related standard deviation obtained for different values of \ac{SR}.
Results show that the \ac{LLE} is not significantly different from zero. In Fig.~\ref{fig:Lyapunov} panel (b), instead, we show the Lyapunov spectrum of a network with $N = 100$ neurons in the recurrent layer, obtained for different \ac{SR} values.
Again, our results show that the maximum \ac{LLE} of \eqref{eqn:new_activation} is zero for finite-size networks, regardless of the values chosen for the \ac{SR}.

\subsection*{Network state dynamics: input-driven case}

Let us define the ``effective input'' contributing to the neuron activation as $\vec{u} := W_{\text{in}} \vec{s}$.
Accordingly, Eq.~\ref{eqn:new_linear_update} takes on the following form:
\begin{equation}
    \vec{a}_{k} = W\vec{x}_{k-1}+ \vec{u}_{k}, \label{eqn:input_3}
\end{equation}
where $\vec{u}_{k}$ operates as a time-dependent bias vector.
Let us define the normalization factor as:
\begin{equation}\label{eqn:norm_factor}
N_k 
= \lVert W \vec{x}_{k-1} + \vec{u}_{k} \rVert,
\end{equation}
Then, as shown in the Methods section, the state at time-step $n$ can be written as:
\begin{equation} \label{eqn:final_state}
    \vec{x}_n = M^{(n, 0)} \vec{x}_0 + \sum^{n}_{k=1} M^{(n, k)} \vec{u}_{k}
\end{equation}
where
\begin{equation}\label{eqn:memory_matrix}
    M^{(n, k)} := \frac{W^{n-k}}{\prod_{l = k}^{n}  N_{l}}.
\end{equation}

By looking at \eqref{eqn:final_state}, it is possible to note that each $\vec{u}_k$ is multiplied by a time-dependent matrix, i.e., the network's final state $\vec{x}_n$ is obtained as a linear combination of all previous inputs. 

\subsection*{Memory of past inputs}

Here, we analyze the ability of the proposed model \eqref{eqn:new_system} to retain memory of past inputs in the state.
In order to perform such analysis, we define a measure of network's memory that quantifies the impact of past inputs on current state $\vec{x}_n$.
Our proposal shares similarities with a memory measure called Fisher memory, first proposed by Ganguli \textit{et al.} \cite{ganguli2008memory} and then further developed in \cite{tivno2013short, tivno2018asymptotic}.
However, our measure can be easily applied also to non-linear networks like the one Eq.~\ref{eqn:new_system} proposed here, justifying the analysis discussed in the following.

Considering one time-step in the past, we have:
\begin{equation}
\label{eqn:stateupdate_one-step}
    \vec{x}_n = \frac{W}{N_n}\vec{x}_{n-1} + \frac{1}{N_n}\vec{u}_{n}
\end{equation}
All past history of the signal is processed in $\vec{x}_{n-1}$.
Note that \eqref{eqn:stateupdate_one-step} keeps the same form for every $n$.
Going backward in time one more step, we obtain:
\begin{equation}\label{eqn:update}
    \vec{x}_n = \frac{W}{N_n}\left(\frac{W}{N_{n-1}}\vec{x}_{n-2} + \frac{1}{N_{n-1}}\vec{u}_{n-1} \right)+ \frac{1}{N_n}\vec{u}_{n}
\end{equation}
As $\lVert \vec{x}_{n} \rVert = \lVert \vec{x}_{n-1} \rVert= \lVert \vec{x}_{n-2} \rVert = 1$, we see that there is a sort of recursive structure in this procedure, where each incoming input is summed to the current state and then their sum is normalized. This is a key feature of the proposed architecture. In fact, it guarantees that each input will have an impact on the state, since the norm of the activations will not be too big or too small, because of the normalization.
We now express this idea in a more formal way.

The norm of \eqref{eqn:memory_matrix} can be written as
\begin{equation}
    \lVert M^{(n, k)} \rVert =
    \frac{1}{\prod_{l=k}^{n}  N_{l}} \lVert W^{n-k} \rVert
    \overset{n-k \gg 1}{\approx} \frac{1}{\prod_{l=k}^{n}  N_{l}} \rho^{n-k}
\end{equation}
where the approximation holds thanks to the Gelfand's formula\footnote{The Gelfand's formula \cite{lax2002functional} states that for any matrix norm $\rho(A)=\lim_{k \to \infty} \left \|A^k \right \|^{\frac{1}{k}}$.}.
%
If the input is null, we expect each $N_l$ to be of the order of $\rho$ as $\lVert x_l \rVert = 1$.
So we write:
\begin{equation}
\label{eqn:impact_input}
N_l = \lVert W \vec{x}_{l-1} + \vec{u}_{l} \rVert 
=\rho + \delta_l
\end{equation}
where $\delta_l$ denotes the impact of the $l$-th input on the $l$-th normalization factor. Its presence is due to the fact that without any input $N_l$ would be approximately equal to $\rho$, while the  input modifies the state and so the normalization value will be modified accordingly.
The value $\delta_l$ is introduced to account for this fact: $\delta_ := N_l - \rho$.

If we assume that each input produces a similar effect on the state (i.e. $\delta_l = \delta$ for every $l$), we finally obtain:
\begin{equation}\label{eqn:module}
    \lVert M^{(n, k)} \rVert 
    \approx
    \frac{\rho^{n-k}}{(\rho+\delta)^{n-k}} 
\end{equation}
We note that such assumption is reasonable for inputs that are symmetrically distributed around a mean value with relatively small variance, e.g. for Gaussian or uniformly distributed inputs (as demonstrated by our results).
However, our argument might not hold for all types of inputs and a more detailed analysis is left as future research.

We define the memory of the network $\mathcal{M}_\alpha(k|n)$ as the influence of input at time $k$ on the network state at time $n$. 
More formally, having defined $\alpha := \rho / \delta$ (since \eqref{eqn:module} only depends on this ratio), we use \eqref{eqn:module} to define the memory as:
\begin{equation}\label{eqn:impact}
    \mathcal{M}_\alpha(k|n) :=
    \left(\frac{\alpha}{\alpha+1}\right)^{n-k}
    =\left(1 - \frac{1}{\alpha + 1}\right)^{n-k}
\end{equation}
Eq. \ref{eqn:impact} goes to $0$ (i.e., no impact of the input on the states) as $\alpha \to 0$ and tends to $1$ for $\alpha\rightarrow\infty$, implying that for an infinitely large \ac{SR} the network perfectly remembers its past inputs, regardless of their occurrence in the past.
Note that \eqref{eqn:impact} does not depend on $k$ and $n$ individually, but only on their difference (elapsed time): the larger the difference, the lower the value of $\mathcal{M}_\alpha(k|n)$, meaning that far away inputs have less impact on the current state.

\subsection*{Memory loss}
By using \eqref{eqn:impact}, we define the memory loss between state $\vec{x}_n$ and $\vec{x}_m$ of the signal at time-step $k$ (with $m>n>k$ and $m = n+ \tau$) as follows:

\begin{equation}
\label{eqn:memory_loss}
\begin{split}
\Delta \mathcal{M}(k|m,n) :=&
\mathcal{M}_\alpha(k|m)  - \mathcal{M}_\alpha(k|n) =
\left(\frac{\alpha}{\alpha+1}\right)^{m-k} - \left(\frac{\alpha}{\alpha+1}\right)^{n-k}
\\
=& \frac{(\alpha+1) ^{k}}{\alpha^{ k}} \frac{\alpha^{ n}}{(\alpha+1) ^{n}} \left(  \frac{\alpha^{\tau}}{(\alpha+1) ^{\tau}} - 1\right) \\
=& \left(1 - \frac{1}{\alpha+1}\right)^n \left(1 + \frac{1}{\alpha}\right)^k \left( \left(1 -\frac{1}{\alpha+1}\right)^{\tau}-1 \right) \le 0
\end{split}
\end{equation}
For very large values of $\alpha$, we have $\Delta \mathcal{M}(k|m,n) \to 0$, meaning that in our model \eqref{eqn:new_activation} larger spectral radii eliminate memory losses of past inputs.
Now, we want to assess if inputs in the far past have higher/lower impact on the state more than recent inputs.
To this end, we selected $n>k>l$ and defined $k = n - a$ and $l = n - b$, leading to $b > a > 0$ and $b = a + \delta$.
Define:
\begin{equation}
\begin{split}
    \Delta \mathcal{M}(k, l|n)  =& \mathcal{M}(n - a|n) - \mathcal{M}(n-a - \delta |n) = \frac{\alpha^{a}}{(\alpha+1)^a} - \frac{\alpha^{a +\delta}}{(\alpha+1)^{a + \delta}} = \\ =
    &\frac{\alpha^{a}}{(\alpha+1)^a}\left(1 - \frac{\alpha^\delta}{(\alpha+1)^\delta}\right) =
    \left(1 - \frac{1}{\alpha + 1}\right)^a \left(1 - \left(1 - \frac{1}{\alpha + 1}\right)^\delta\right) \ge 0
\end{split}
\end{equation}

We have that $ \lim_{\delta \to \infty} \Delta \mathcal{M}(n - a, n - a - \delta |n) = \left(1 - \frac{1}{\alpha + 1}\right)^a$, showing how the impact of an input that is infinitely far in the past is not null compared with one that is only $a<\infty$ steps behind the current state. This implies that the proposed network is able to preserve in the network state (partial) information from all inputs observed so far.

\subsubsection*{Linear networks} In order to assess the memory of linear models, we perform the same analysis for linear \acp{RNN} (i.e., $\vec{x}_n = \vec{a}_n$) by using the definitions given in the previous section.
In this case, an expression for the memory can be obtained straightforwardly and reads:
\begin{equation}
\label{eqn:memory_linear}
     \mathcal{M}_L(k|n) := \rho^{n-k}
\end{equation}
It is worth noting that there is no dependency on $\delta$ and, therefore, on the input.
Just like before, we have the memory loss of the signal at time-step $k$ between state $\vec{x}_n$ and $\vec{x}_m$, as:
\begin{align}\label{eqn:memory_linear_fixed_step}
     \Delta \mathcal{M}_L(k|m,n) &=\mathcal{M}_L(k|m) - \mathcal{M}_L(k|n)
    = \mathcal{M}_L(k|n+\tau) - \mathcal{M}_L(k|n)
     = 
     (\rho^\tau -1) \rho^{n-k} \le 0
\end{align}
where we set $m>n>k$ and $m = n+ \tau$.
As before, we also discuss the case of two different inputs observed before time-step $n$.
By selecting time instances $n>k>l$, $k = n - a$ and $l = n - b$, we have $b > a > 0$ allowing us to write $b = a + \delta$.
As such:
\begin{align}\label{eqn:memory_linear_fixed_state}
     \Delta \mathcal{M}_L(k,l|n) &=\mathcal{M}_L(k|n) - \mathcal{M}_L(l|n)
     = 
     (1 - \rho^\delta ) \rho^{a} \ge 0.
\end{align}
We see that in both \eqref{eqn:memory_linear_fixed_step} and \eqref{eqn:memory_linear_fixed_state} the memory loss tends to zero as the spectral radius tends to one, which is the critical point for linear systems.
So, according to our analysis, linear networks should maximize their ability to remember past inputs when their $\ac{SR}$ in close to one and, moreover, their memory should be the same disregarding the particular signal they are dealing with. 
We will see in the next section that both these claims are confirmed by our simulations.

\subsubsection*{Hyperbolic tangent} 
We now extend the analysis to standard \acp{ESN}, i.e., those using a $\tanh$ activation function. Define $\phi:=\tanh$ (applied element-wise). Then, for generic scalars $a$ and $b$, when $|b| \ll 1$ the following approximation holds:
\begin{equation}
\phi(\vec{a}+\vec{b}) = \frac{\phi(\vec{a}) + \phi(\vec{b})}{1 + \phi(\vec{a}) \phi(\vec{b})} \approx \phi(\vec{a}) + \phi(\vec{b})
\end{equation}
When $\vec{a}$ is the state and $\vec{b}$ is the input, our approximation can be understood as a small-input approximation \cite{verstraeten2009quantification}.
Then, the state-update reads:
\begin{equation}
\begin{split}
    \vec{x}_n &= \phi(W \vec{x}_{n-1} + \vec{u}_n) \approx \phi(W \vec{x}_{n-1}) + \phi(\vec{u}_n) \\
    &= \phi(W\phi(W \vec{x}_{n-2} + u_{n-1})) + \phi(\vec{u}_n) 
    \approx
    \phi(W\phi(W \vec{x}_{n-2})) +\phi(W \phi(\vec{u}_{n-1}))+ \phi(\vec{u}_n)
\end{split}
\end{equation}
Thus, applying the same argument used in the previous cases, it is possible to write:
\begin{align}\label{eqn:tanh_memory}
    \lVert \vec{x}_n \rVert  
    &=  S_{\rho}^{(0)}(\phi(\lVert \vec{u}_n) \rVert)+S_{\rho}^{(1)}(\phi(\lVert \vec{u}_{n-1}\rVert)) + \dots + S_{\rho}^{(n)}(\phi(\lVert \vec{u}_0\rVert))
    =
    \sum^n_{k=0} S_{\rho}^{(n-k)}(\phi(\lVert \vec{u}_{k}\rVert))
    \end{align}
where we defined:
\begin{equation}
    S_{\rho}^{(n)}(\xi) := \underbrace{\phi(\rho \cdot {\dots}\phi (\rho\cdot \phi(}_{n \text{ times}} \xi)))
\end{equation}
\\
We see that, differently from the previous cases, the final state $\vec{x}_{n}$ is a sum of nonlinear functions of the signal $\vec{u}_k$.
Each signal element $\vec{u}_{k}$ is encoded in the network state by first multiplying it by $\rho$ and then passing the outcome through the nonlinearity $\phi(\cdot)$.
This implies that, for networks equipped with hyperbolic tangent function, larger spectral radii favour the forgetting of inputs (as we are in the non-linear region of $\phi(\cdot)$).
On the other hand, for small spectral radii the network operates in the linear regime of the hyperbolic tangent and the network behaves like in the linear case.

\begin{figure}[ht!]
\centering
\includegraphics[width=0.5\linewidth]{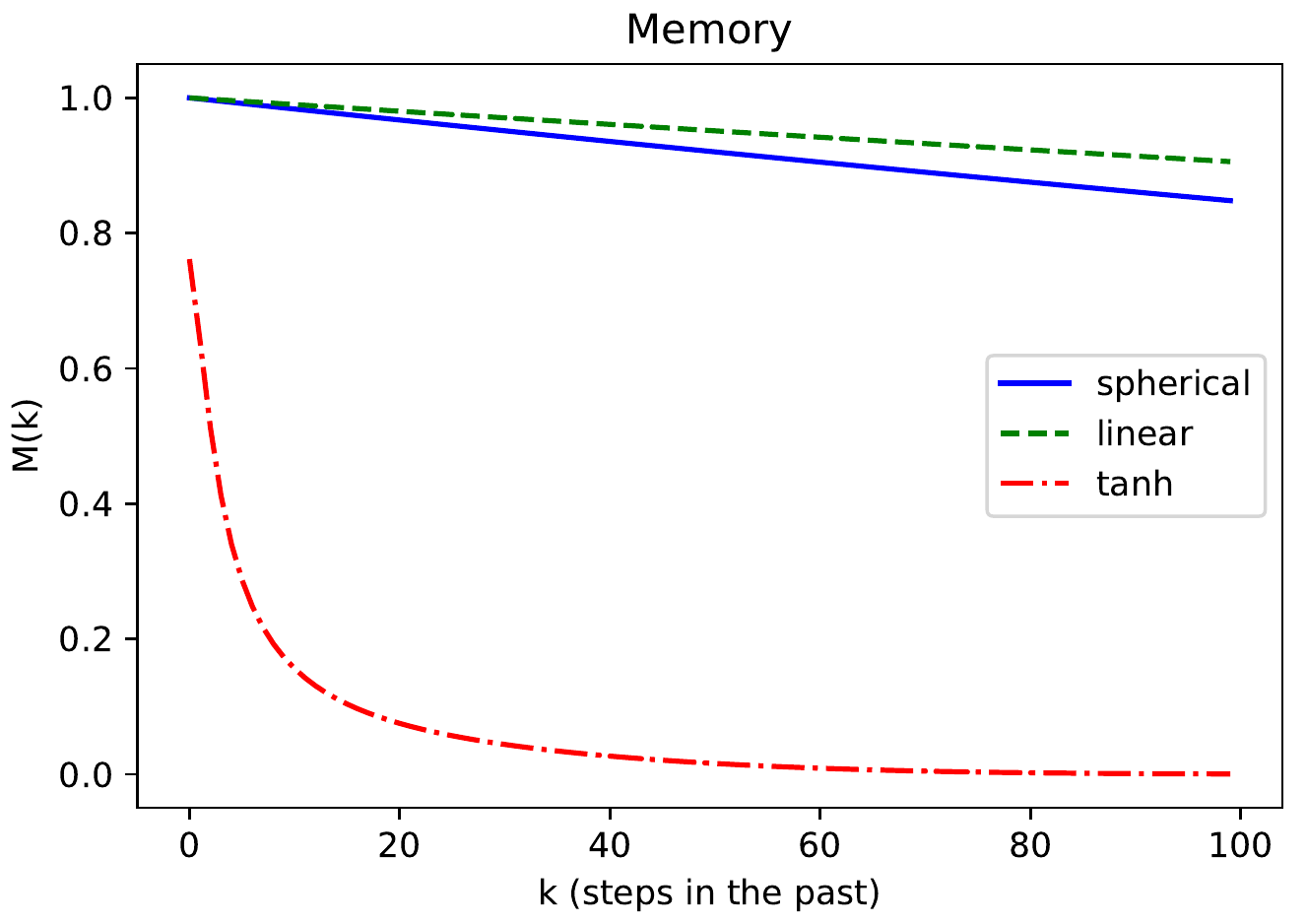}
\caption{Decay of the memory in the cases associated with expressions \eqref{eqn:impact}, \eqref{eqn:memory_linear}, and \eqref{eqn:tanh_memory}. In all cases, the decay is consistent with the predictions. The memory for the hyperbolic tangent is estimated assuming a signal of magnitude $1$.}
\label{fig:memory-theor}
\end{figure}

A plot depicting the decay of the memory of input at time-step $k$ for all three cases described above is shown in Fig.~\ref{fig:memory-theor}.
The trends demonstrate that, in all cases, the decay is consistent with our theoretical predictions.

\section*{Performance on memory tasks}
We now perform experiments to assess the ability of the proposed model to reproduce inputs seen in the past, meaning that the desired output at time-step $k$ is $\vec{y}_k = \vec{u}_{k-\tau}$, where $\tau$ ranges from $0$ to $100$. We compare our model with a linear network and a network with hyperbolic tangent ($\tanh$) activation functions.
We use fixed, but reasonable hyper-parameters for all networks, since in this case we are only interested in analyzing the network behavior on different tasks. In particular, we selected hyper-parameters that worked well in all cases taken into account; in preliminary experiments, we noted that different values did not result in substantial (qualitative) changes of the results.
The number of neuron $N$ is fixed to $1000$ for all models.
For linear and nonlinear networks, the input scaling (a constant scaling factor of the input signal) is fixed to $1$ and the \ac{SR} equals $\rho = 0.95$. 
For the proposed model \eqref{eqn:new_system}, the input scaling is chosen to be $0.01$, while the \ac{SR} is $\rho = 15$. 
For the sake of simplicity, in what follows we refer to \acp{ESN} resulting from the use of \eqref{eqn:new_system} as ``spherical reservoir''.

To evaluate the performance of the network, we use the accuracy metric defined as $\gamma = \max\{1 - \text{NRMSE}, 0\}$, where the \ac{NRMSE} is defined as:
\begin{equation}\label{eqn:error}
    \text{NRMSE} := \sqrt{\frac{ \langle \lVert\vec{y}_k - \vec{\hat{y}}_k\lVert^2 \rangle}{ \langle \lVert \vec{y}_k - \langle \vec{y}_k \rangle \rVert^2 \rangle}}.
\end{equation}
Here, $\vec{\hat{y}}_k$ denotes the computed output and $\langle \cdot \rangle$ represents the time average over all time-steps taken into account.
In the following, we report the accuracy $\gamma$ while varying $\tau$ in the past for various benchmark tasks.
Each result in the plot is computed using the average over $20$ runs with different network initializations. 
The networks are trained on a data set of length $L_\text{train}=5000$ and the associated performance is evaluated using a test set of length $L_\text{test}=2000$.
The shaded area represents the standard deviation. All the considered signals were normalized to have unit variance.
\begin{figure}
    \centering
    \includegraphics[width=\linewidth]{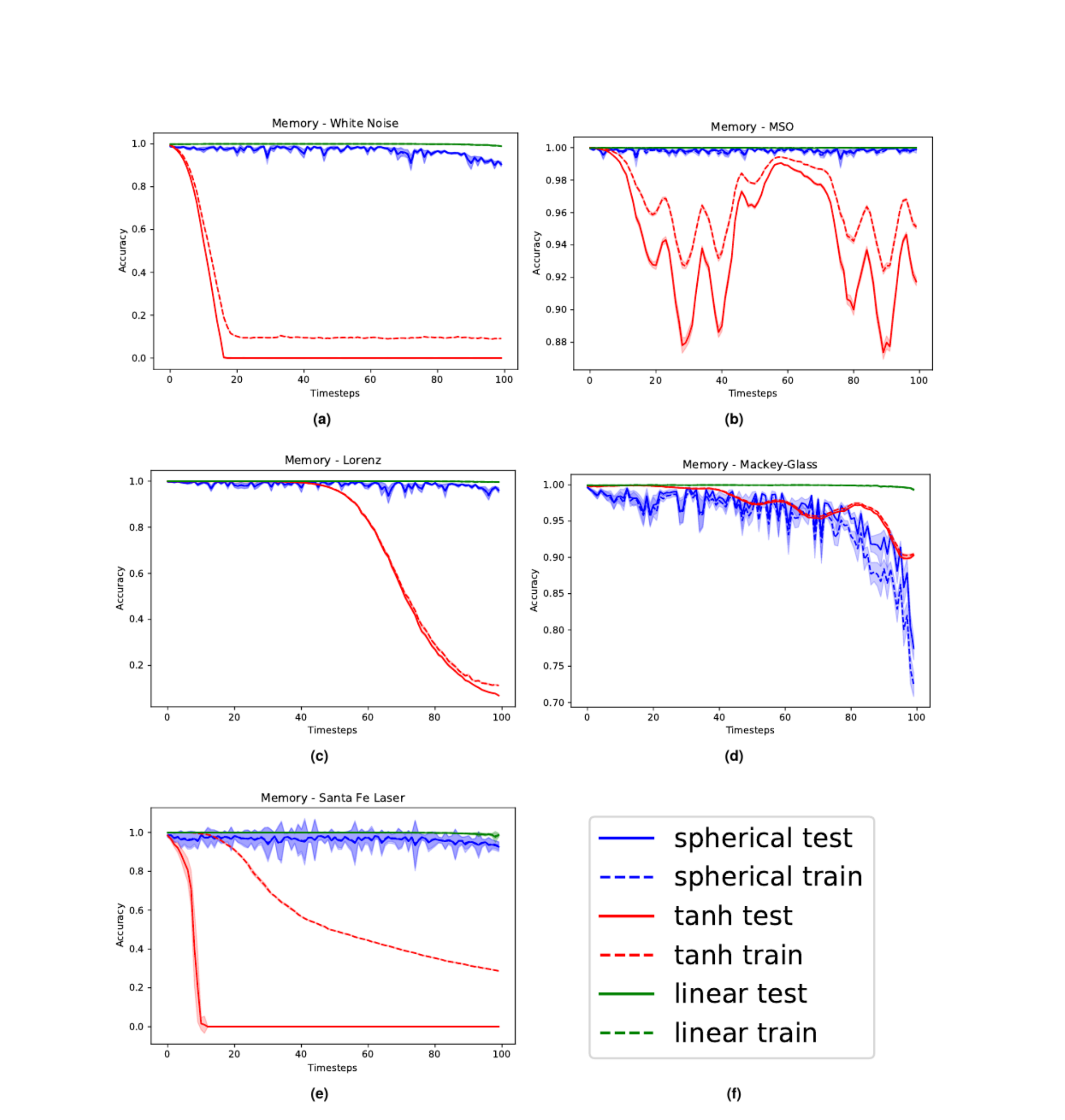} 
    
    \caption{
    Results of the experiments on memory for different benchmarks. Panel (a) displays the white noise memorization task, (b) the \ac{MSO}, (c) the $x$-coordinate of the Lorenz system, (d) the Mackey-Glass series and (e) the Santa Fe laser dataset. 
    As described in the legend (f), different line types account for results obtained on training and test data. The shaded areas represent the standard deviations, computed using $20$ different realization for each point.}
    \label{fig:memory}
\end{figure}

\subsubsection*{White noise}
In this task, the network is fed with white noise uniformly distributed in the $[-1,1]$ interval. Results are shown in Fig.\ref{fig:memory}, panel (a). We note that networks using the spherical reservoir have a performance comparable with linear networks, while $\tanh$ networks do not correctly reconstruct the signal when $k$ exceeds $20$.
It is worth highlighting the similarity of the plots shown here with our theoretical predictions about the memory (Eqs. \eqref{eqn:impact}, \eqref{eqn:memory_linear}, and \eqref{eqn:tanh_memory}) shown in Fig.~\ref{fig:memory-theor}.

\subsubsection*{Multiple superimposed oscillators}
The network is fed with Multiple Superimposed Oscillators \ac{MSO} with $3$ incommensurable frequencies:
\begin{equation}\label{eqn:MSO}
    u(k) = \sin(0.2 k) + \sin (0.311 k) + \sin (0.42 k)
\end{equation}
Results are shown in Fig.\ref{fig:memory}, panel (b). This task is difficult because of the multiple time scales characterizing the inputs\cite{jaeger2004harnessing}.
We note the performance of the linear and the spherical reservoirs are again similar and both outperform the network using the hyperbolic tangent.
The accuracy peak when $k \approx 60$ is due to the fact that the autocorrelation of the signal reaches its maximum value at that time-step.

\subsubsection*{Lorenz series}
    The Lorenz system is a popular mathematical model, originally developed to describe atmospheric convection.
    \begin{equation}
    \begin{cases}
        &\dot{x}= \sigma (y - x), \\
        &\dot{y}= x (\rho - z) - y, \\
        &\dot{z}= x y - \beta z.
        \end{cases}
    \end{equation}
    It is well-known that this system is chaotic when $\sigma = 10$, $\beta = \frac{8}{3}$ and $\rho = 28$.
    We simulated the system with these parameters and then fed the network with the $x$ coordinate only.
    Results are shown in Fig.\ref{fig:memory}, panel (c). Also in this case, while the accuracy for spherical and linear networks does not seem to be affected by $k$, the performance of networks using the $\tanh$ activation dramatically decreases when $k$ is large. This stresses the fact that non-linear networks are significantly penalized when they are requested to memorize past inputs.

\subsubsection*{Mackley-Glass system}
    The Mackley-Glass system is given by the following delayed differential equation:
    \begin{equation} \label{eqn:MG}
        \dot{x}(t) = -\beta x(t) + \frac{\alpha x(t - \lambda)}{ 1 + x(t - \lambda)} 
    \end{equation}
    In our experiments, we simulated the system using standard parameters, that is, $\lambda = 17$, $\alpha = 0.2$ and $\beta = 0.1$. 
    Results are shown in Fig.\ref{fig:memory}, panel (d). Note that in this case the performance of the network with spherical reservoir is comparable with the one obtained using the hyperbolic tangent and both of them are outperformed by the linear networks.
    
\subsubsection*{Santa Fe laser dataset}
    The Santa Fe laser dataset is a chaotic time series obtained from laser experiments.
    The results are shown in Fig.\ref{fig:memory}, panel (e). Also in this case the hyperbolic tangent networks do not manage to remember the signal, while the other systems show the usual behavior.

\section*{Memory/non-linearity trade-off}
Here, we evaluate the capability of the network to deal with tasks characterized by tunable memory and non-linearity features \cite{inubushi2017reservoir}. 
The network is fed with a signal where the $u_k$s are univariate random variables drawn from a uniform distribution in $[-1,1]$. The network is then trained to reconstruct the desired output of $y_k = \sin(\nu \cdot u_{k-\tau})$. We see that $\tau$ accounts for the memory required to correctly solve the task, while $\nu$ controls the amount non-linearity involved in the computation.
For each configuration of $\tau$ and $\nu$ chosen in suitable ranges, we run a grid search on the range of admissible values of \ac{SR} and input scaling. Notably, we considered $20$ equally-spaced values of the \ac{SR} and for the input scaling.
For networks using hyperbolic tangent, the \ac{SR} varies in $[0.2,3]$; $[0.2,1.5]$ for linear networks, and $[0.2,10]$ for networks with spherical reservoir. The input scaling always ranges in $[0.01,2]$.
\footnote{This choice of exploring large values of \ac{SR} for the hyperspherical case is motivated by the fact that condition \eqref{eqn:condition} must be satisfied in order for the network to work properly: choosing a large \ac{SR} will also lead to larger $\sigma_\text{min}(W)$, as discussed in the `Universal function approximation property' section.}
We then choose the hyper-parameters configuration that minimizes \eqref{eqn:error} on a training set of length $L_\text{train}=500$ and then assess the error on a test set with $L_\text{test}=200$.

\begin{figure}[ht]
\centering
\includegraphics[keepaspectratio=true,scale=0.5]{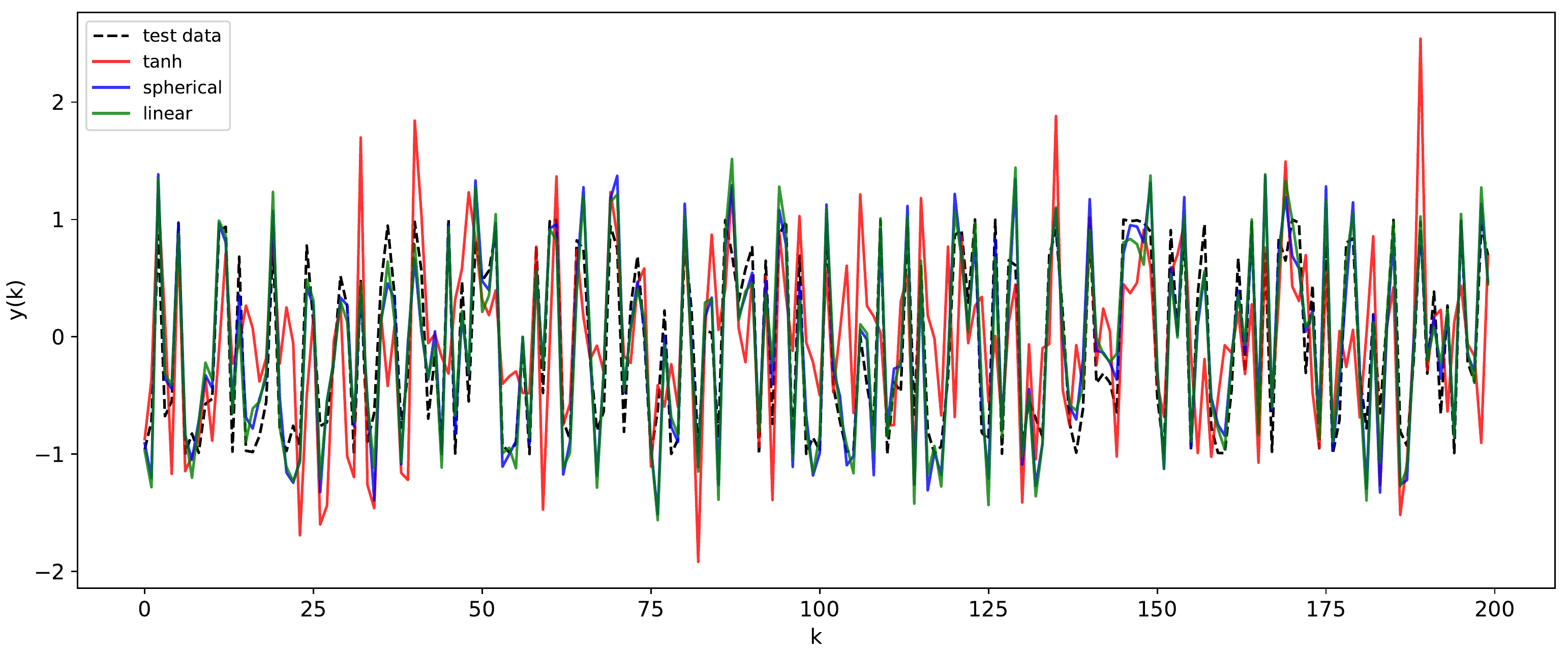}
\caption{Comparison of the network prediction the memory--nonlinearity task for $\nu = 2.5$ and $\tau = 10$. The hyper-parameters of the networks are the same used to generate Fig.~\ref{fig:memory}. Here the accuracy values are 
$\gamma_{\text{tanh}}=0.12$, 
$\gamma_{\text{spherical}}=0.63$ and 
$\gamma_{\text{linear}}=0.61$.}
\label{fig:example}
\end{figure}

\begin{figure}[ht!]
\centering
\includegraphics[width=\linewidth]{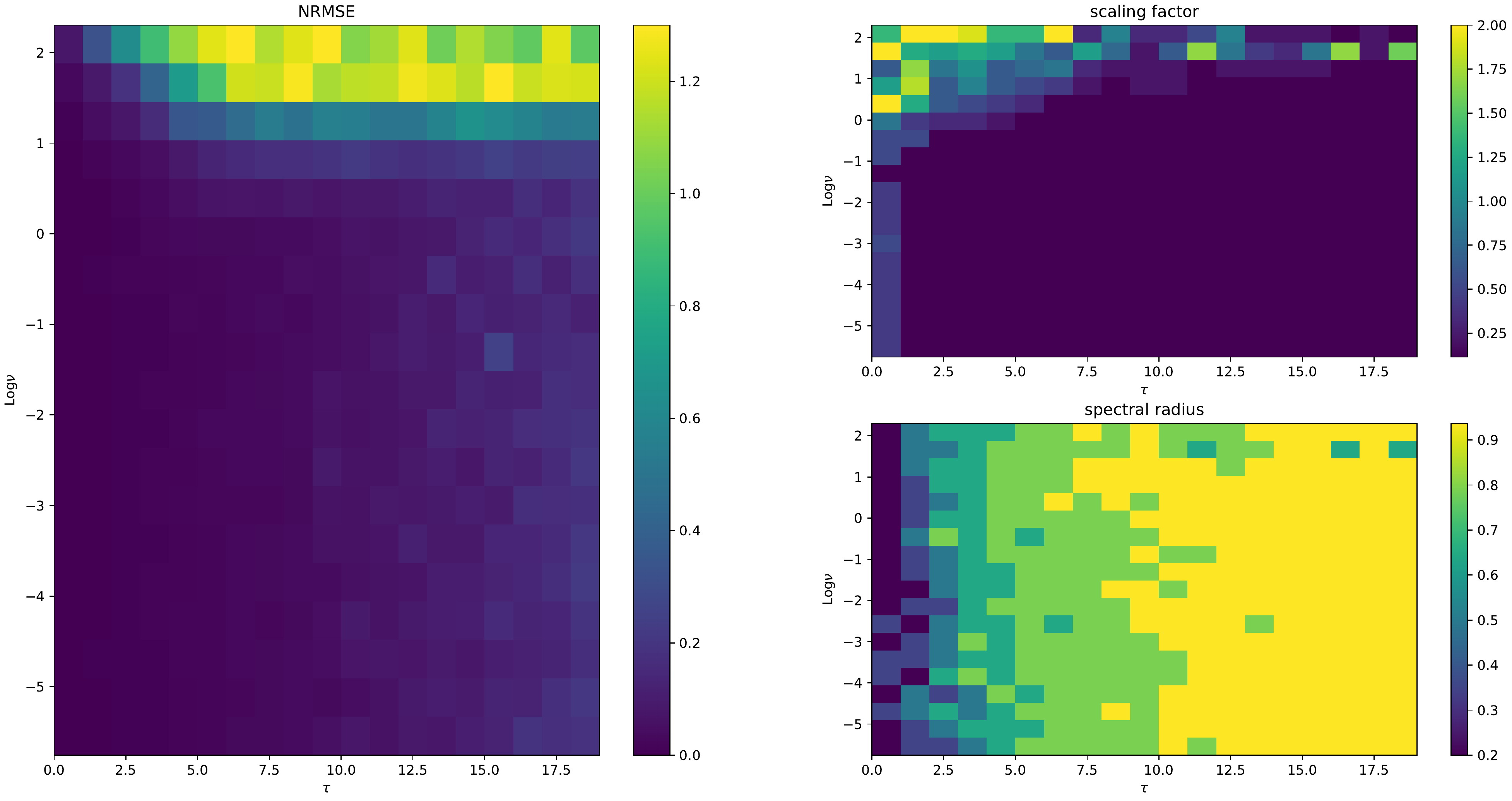}
\caption{Results for the hyperbolic tangent activation function. The network performs as expected: the error grows with the memory required to solve the task. The choice of the spectral radius displays a pattern, where larger \acp{SR} are preferred when more memory is required. The scaling factor tends to be small for almost every configuration.}
\label{fig:regular}
\end{figure}

\begin{figure}[ht!]
\centering
\includegraphics[width=\linewidth]{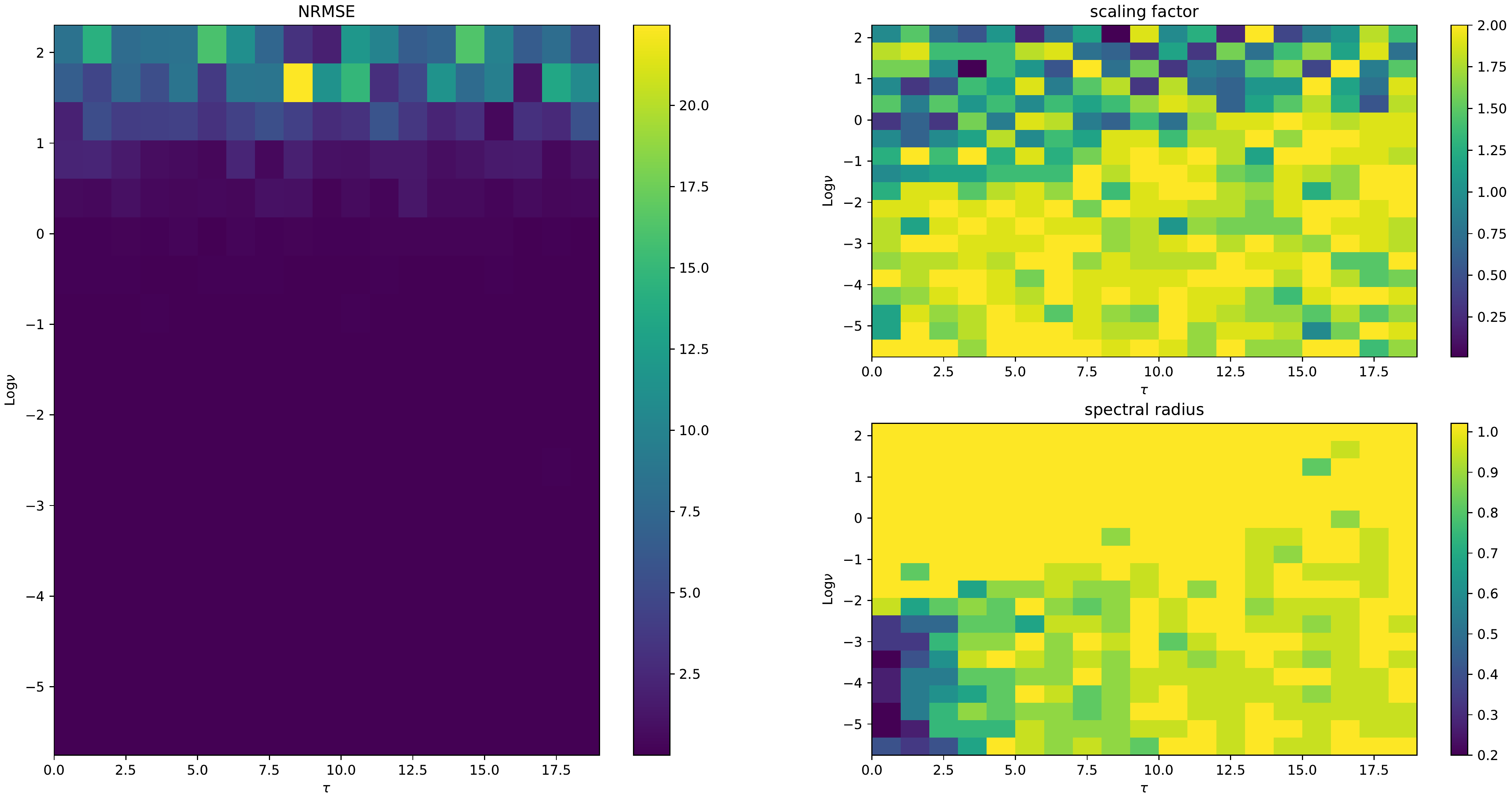}
\caption{Performance of linear networks. We note that $\tau$ seems to have no significant effect on the performance. In fact, we note very large errors when the nonlinearity of the task is high. The choice of the scaling factor and of the spectral radius shows a really weak tendency to certain values, indicating that the performance is only weakly influenced by the hyper-parameters.}
\label{fig:linear}
\end{figure}

\begin{figure}[ht]
\centering
\includegraphics[width=\linewidth]{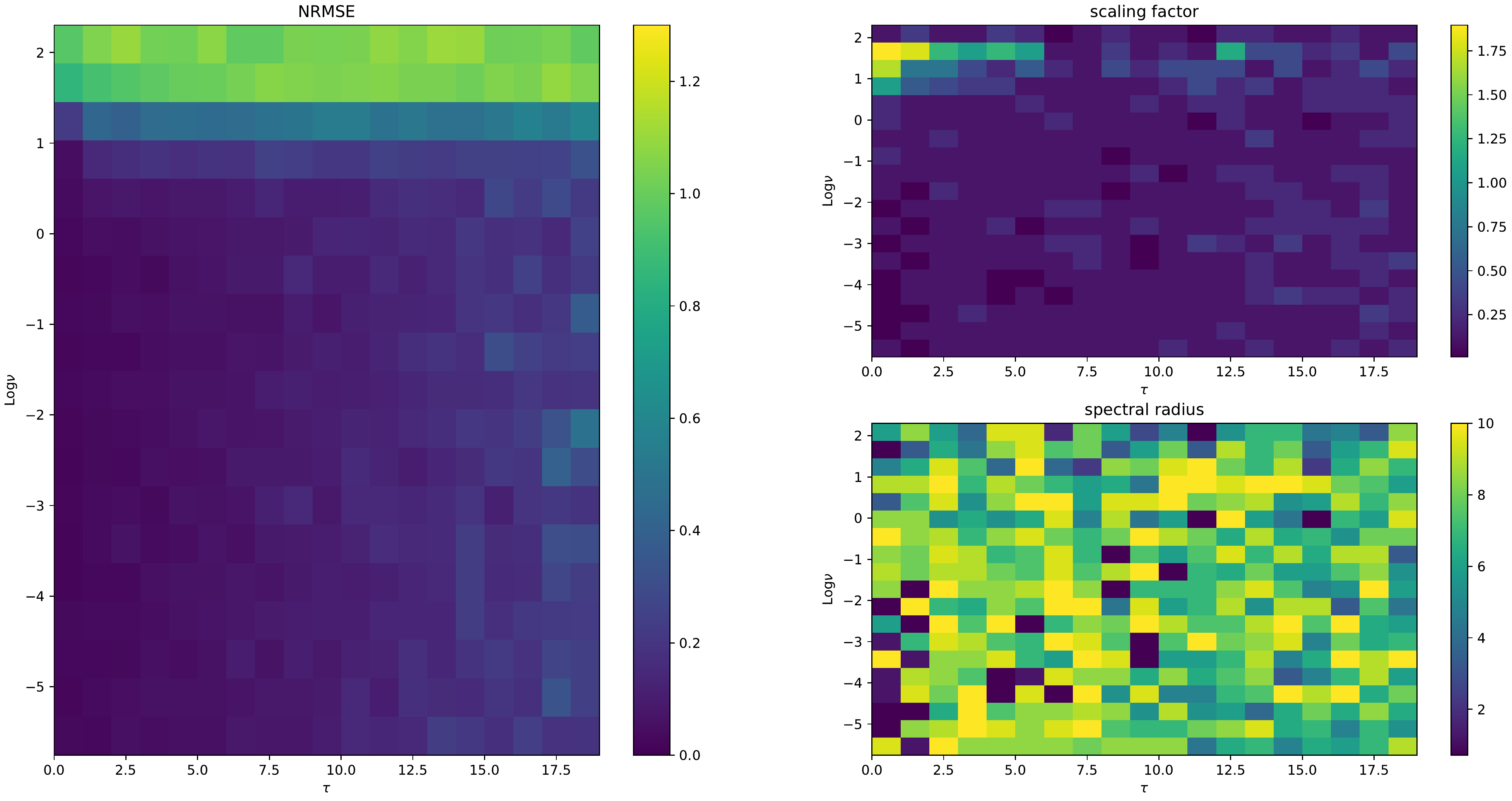}
\caption{Performance of the proposed model. We see that the network performs reasonably well for all the tasks displaying only a weak dependency on the memory. Moreover the spectral radius does not to play any role in the network performance. The choice of the scaling factor denotes similar patterns with the hyperbolic tangent case.}
\label{fig:spherical}
\end{figure}

In Figures \ref{fig:regular}, \ref{fig:linear}, and \ref{fig:spherical}, we show the \ac{NRMSE} for the task described above for different values of $\nu$ and $\tau$, and the ranges of the input scaling factor and the spectral radius which performed best for the hyperbolic tangent, linear and spherical activation function, respectively. 
The results of our simulations agree with those reported in \cite{inubushi2017reservoir} and, most importantly, confirm our theoretical prediction: the proposed model possess memory of past inputs that is comparable with the one of linear networks but, at the same time, it is also able to perform nonlinear computations.
This explains why the proposed model denotes the best performance when the task requires both memory and nonlinearity, i.e., when both $\tau$ and $\nu$ are large.
Predictions obtained for a specific instance of this task requiring both features are given in Fig~\ref{fig:example}, showing how the proposed model outperforms the competitors.
In order to explore the memory-nonlinearity relationship we followed the experimental design proposed in \cite{inubushi2017reservoir}: our goal is to study the memory property of the proposed model and not to develop a model specifically designed to maximize the compromise between memory and nonlinearity. The reader interested in models that aim at explicitly tackling the memory--nonlinearity problem may refer to recently-developed hybrid systems \cite{inubushi2017reservoir, di2018combining} which try to deal with the memory--nonlinearity trade--off by combining, in different ways, linear and non-linear units so that the network can exploit a combination of them according to the problem at hand. These approaches introduce new hyper-parameters which, basically, allow to control the memory--nonlinearity trade--off. Future works will investigate such a trade--off. 

\clearpage
\section*{Discussion}

In this work, we studied the properties of \acp{ESN} equipped with an activation function that projects at each time-step the network state on a hyper-sphere.
The proposed activation function is global, contrarily to most activation functions used in the literature, meaning that the activation value of a neuron depends on the activations of all other neurons in the recurrent layer.
Here, we first proved that the proposed model is a universal approximator just like regular \acp{ESN} and gave sufficient conditions to support this claim.
Our theoretical analysis shows that the behavior of the resulting network is never chaotic, regardless of the setting of the main hyper-parameters affecting its behavior in phase space.
This claim was supported both analytically and experimentally by showing that the maximun Lyapunov exponent is always zero, implying that the proposed model always operates on the \acp{EoC}.
This leads to networks which are (globally) stable for any hyper-parameter configurations, regardless of the particular input signal driving the network, hence solving stability problems that might affect \acp{ESN}.

The proposed activation function allows the model to display a memory capacity comparable with the one of linear \acp{ESN}, but its intrinsic nonlinearity makes it capable to deal with tasks for which rich dynamics are required.
To support this claim, we developed a novel theoretical framework to analyze the memory of the network. By taking inspiration from the Fisher memory proposed by Ganguli \textit{et al.} \cite{ganguli2008memory}, we focused on quantifying how much past inputs are encoded in future network states. The developed theoretical framework allowed us to account for the nonlinearity of the proposed model and predicted a memory capacity comparable with the one of linear \acp{ESN}. This theoretical prediction was validated experimentally by studying the behavior of the memory loss using white noise as inputs as well as well-known benchmarks used in the literature.
Finally, we experimentally verified that the proposed model offers an optimal balance of nonlinearity and memory, showing that it outperforms both linear and \acp{ESN} with $\tanh$ activation functions in those tasks where both features are required at the same time.

\section*{Methods}

\subsection*{Contractivity condition}\label{subsec:contractivity}


%

The universal approximation property, as exhaustively discussed in \cite{grigoryeva2018echo}, can be proved for an \ac{ESN} of the form \eqref{eqn:activation} provided that it has the \ac{ESP}. 
To prove the \ac{ESP}, it is sufficient to show that the network has the contractivity condition, i.e., that for each $\vec{x}$ and $\vec{y}$ in the domain of the activation function $\phi$, it must be true that:
\begin{equation}\label{eqn:contractivity}
   \lVert \phi(\vec{x}) - \phi(y) \rVert \le \lVert \vec{x} - \vec{y} \rVert
\end{equation}
Let us introduce the the notation $\vec{\hat{x}} :=\vec{x}/ \lVert \vec{x} \rVert$ and $\vec{\hat{y}} :=\vec{y}/ \lVert \vec{y} \rVert$. 
Taking the square of the norm, one gets:
\begin{equation}
    \lVert \phi({\vec{x}}) - \phi({\vec{y}})  \rVert^2 = 
  r^2 ( 2 - 2 (\vec{\hat{x}}\cdot \vec{\hat{y}})) 
    \le r^2 \left( \left( \frac{\lVert \vec{x} \rVert}{\lVert \vec{y} \rVert} + \frac{\lVert \vec{y} \rVert}{\lVert \vec{x} \rVert}\right) - 2 \vec{\hat{x}}\cdot \vec{\hat{y}}\right)
\end{equation}

Where $\vec{a} \cdot \vec{b}$ is the scalar product between two vectors and the inequality $\frac{a}{b} + \frac{b}{a} \ge 2$ follows from the fact that $(a-b)^2 = a^2 + b^2 - 2ab > 0 $ for all $a,b > 0$.
Now, we assume $\lVert \vec{x} \rVert, \lVert \vec{y} \rVert > r$ and show that:
\begin{align}
    & 
    r^2 \cdot \left(\frac{\lVert \vec{x} \rVert}{\lVert \vec{y} \rVert} + \frac{\lVert \vec{y} \rVert}{\lVert \vec{x} \rVert} - 2 \vec{\hat{x}}\cdot \vec{\hat{y}} \right)
    \le
    \lVert \vec{x} \rVert \lVert \vec{y} \rVert \left(\frac{\lVert \vec{x} \rVert}{\lVert \vec{y} \rVert} + \frac{\lVert \vec{y} \rVert}{\lVert \vec{x} \rVert} - 2 \vec{\hat{x}}\cdot \vec{\hat{y}} \right) = \lVert \vec{x} \rVert ^2 + \lVert \vec{y} \rVert^2 - \underbrace{2(\lVert \vec{x} \rVert) (\lVert \vec{y} \rVert) \vec{\hat{x}}\cdot \vec{\hat{y}}}_ {2\vec{x}\cdot \vec{y}} =
    \lVert {\vec{x}} - {\vec{y}}  \rVert_2^2
\end{align}
proving the contractivity condition \eqref{eqn:contractivity}.

We see that the only condition needed is $\lVert \vec{x} \rVert, \lVert \vec{y} \rVert > r$, which means that the linear part of the update \eqref{eqn:new_linear_update} must map states outside the hyper-sphere of radius $r$.
Finally, by applying properties of norms, we observe that:
\begin{equation}
    \lVert {W\cdot\vec{x}_k+ W_{in}\cdot\vec{s}_k} \rVert 
    \ge
    \lVert {W\cdot\vec{x}_k \rVert - \lVert W_{in}\cdot\vec{s}_k} \rVert
    \ge
    \sigma_\textit{min}(W)r- \lVert W_{in} \rVert \lVert\vec{s}_{\textit{max}}\rVert
\end{equation}
and asking this to be larger than $r$ results in the condition \eqref{eqn:condition}.

\subsection*{Input-driven dynamics}\label{subsec:input_driven}

Here we show how to derive \eqref{eqn:final_state}.
Consider a dynamical system evolving according to \eqref{eqn:new_system}: if we explicitly expand the first steps from the initial state $\vec{x}_0$ we obtain:
\begin{align}
x_{1} &= \frac{W\vec{x_0}+\vec{u}_{1}}{\lVert W \vec{x_0}+\vec{u}_{1} \rVert} = \frac{W\vec{x_0}+\vec{u}_{1}}{N_1}=
\frac{W}{N_1}\vec{x}_0 + \frac{1}{N_1}\vec{u}_{1}
\\
x_{2} &= \frac{W\vec{x_1}+\vec{u}_{1}}{N_2} = \frac{W^2}{N_2 N_1}\vec{x}_0 + \frac{W }{N_2 N_1}\vec{u}_{1} + \frac{1}{N_2}\vec{u}_{2}\\
x_{3} &= \frac{W\vec{x_2}+\vec{u}_{3}}{N_3} = \frac{W^3}{N_3 N_2 N_1}\vec{x}_0 + \frac{W^2}{N_3 N_2 N_1}\vec{u}_{1} + \frac{W}{N_3 N_2}\vec{u}_{2} + \frac{1}{N_3}\vec{u}_{3}
\end{align}
So that the general case can be written as:
\begin{align}
x_{n} &= \frac{W\vec{x}_{n-1}+\vec{u}_{n}}{N_n} = \frac{W^{n}}{N_n N_{n-1} \dots N_1}\vec{x}_0 + \frac{W^{n-1}}{N_n N_{n-1} \dots N_1}\vec{u}_{1} + \frac{W^{n-2}}{N_n N_{n-1} \dots N_2}\vec{u}_{2}  
+ \dots + \frac{1}{N_n}\vec{u}_{n}
\end{align}
and in a more compact form as follows:
\begin{equation}
    x_n = M^{(n, 0)} \vec{x}_0 + \sum^{n}_{k=1} M^{(n, k)} \vec{u}_{k}
\end{equation}
where
\begin{equation}
    M^{(n, k)} := \frac{W^{n-k}}{\prod_{l = k}^{n}  N_{l}}
\end{equation}
and $N_0 = 1$.

\bibliography{sample}

\section*{Acknowledgements}

We gratefully acknowledge the support of NVIDIA Corporation with the donation of the Titan Xp GPU used for this research.
LL gratefully acknowledges partial support of the Canada Research Chairs program.

\section*{Author contributions statement}
L.L. outlined the research ideas. P.V. conceived the methods and performed the experiments. C.A. contributed to the technical discussion. All authors developed the theoretical framework, took part to the paper writing and approved the final manuscript.

\section*{Additional information}
\textbf{Competing interests} The authors declare no competing  interests.


%
%
\begin{acronym}
    \acro{bESN}{binary ESN}
	\acro{EoC}{Edge of Criticality}
	\acro{ESN}{Echo State Network}
	\acro{ESP}{Echo State Property}
	\acro{FIM}{Fisher Information Matrix}
	\acro{LLE}{Local Lyapunov Exponent}
	\acro{LSM}{Liquid State Machine}
	\acro{MFT}{Mean Field Theory}
	\acro{ML}{Machine Learning}
	\acro{MSO}{Multiple Superimposed Oscillator}
	\acro{NRMSE}{Normalized Root Mean Squared Error}
	\acro{PDF}{Probability Density Function}
	\acro{RBN}{Random Boolean Network}
	\acro{RNN}{Recurrent Neural Network}
	\acro{RP}{Recurrency Plot}
	\acro{SR}{Spectral Radius}
\end{acronym}

\end{document}